\documentclass{article}

     \usepackage[final,nonatbib]{neurips_2019}

\usepackage[utf8]{inputenc} 
\usepackage[T1]{fontenc}    
\usepackage{hyperref}       
\usepackage{url}            
\usepackage{booktabs}       
\usepackage{amsfonts}       
\usepackage{nicefrac}       
\usepackage{microtype}      
\usepackage{float}
\usepackage{graphicx}
\usepackage{subcaption}
\usepackage{hyperref}
\usepackage{authblk}
\usepackage{amsmath}
\usepackage[numbers,sort]{natbib}
\usepackage{mathtools}

\title{High- and Low-level image component decomposition using VAEs for improved reconstruction and anomaly detection}

\author{
  \textbf{David Zimmerer, Jens Petersen, Klaus Maier-Hein}\\
  Division of Medical Image Computing\\
  German Cancer Research Center (DKFZ)\\
  Heidelberg, Germany \\
}

\begin{document}

\maketitle

\begin{abstract}
  Variational Auto-Encoders have often been used for unsupervised pretraining, feature extraction and out-of-distribution and anomaly detection in the medical field. However, VAEs often lack the ability to produce sharp images and learn high-level features. We propose to alleviate these issues by adding a new branch to conditional hierarchical VAEs. This enforces a division between higher-level and lower-level features. Despite the additional computational overhead compared to a normal VAE it results in sharper and better reconstructions and can capture the data distribution similarly well (indicated by a similar or slightly better OoD detection performance).
\end{abstract}

\section{Introduction}
In medical imaging Variational Auto-Encoders (VAEs) have often been used for unsupervised pretraining, feature extraction and out-of-distribution (OoD) / anomaly detection \cite{baur_deep_2018, chen_deep_2018, zimmerer_case_2018, litjens_survey_nodate, shin_stacked_nodate}. However, VAEs are often criticized to have over-smoothed reconstructions and only learn low-level statistics \cite{nalisnick_deep_2018, larsen_autoencoding_2016, dai_diagnosing_2019, razavi_generating_2019}. Especially in medical applications a faithful reconstruction and small textural differences can be important. As a solution, hierarchical models have been proposed \cite{maaloe_biva:_2019, sonderby_ladder_2016, zhao_learning_2017}. While hierarchical models are very successful in (medical) segmentation with the U-Net \cite{ronneberger_u-net:_2015} as a prime example, research has shown that for unsupervised learning these hierarchical models currently show only a small or no improvement compared to basic VAEs when learning and dividing between high- and low-level features \cite{zhao_learning_2017, maaloe_biva:_2019}. One example is the Ladder VAE \cite{sonderby_ladder_2016} which in practice has been shown to collapse to a single level model \cite{zhao_learning_2017}. BIVA \cite{maaloe_biva:_2019} was shown to give some improvements by using a very deep multi-stage architecture where each level is conditioned on the lower and higher levels. 

Here, inspired by PCA and VQVAE-2 \cite{razavi_generating_2019}, we introduce a simpler hierarchical model with a low-level reconstruction branch which enforces the partitioning into higher-level and low-level components which in our case correspond to the coarser and the finer structure of brain MRIs. The model, which we call primary components conditional hierarchical VAE (pchVAE), shows better reconstructions than a normal VAE while having similar or slightly better OoD detection performance (indicating they capture the data distribution similarly well).  

\section{Methods}

\begin{figure}
  \centering
    \includegraphics[width=0.9\textwidth]{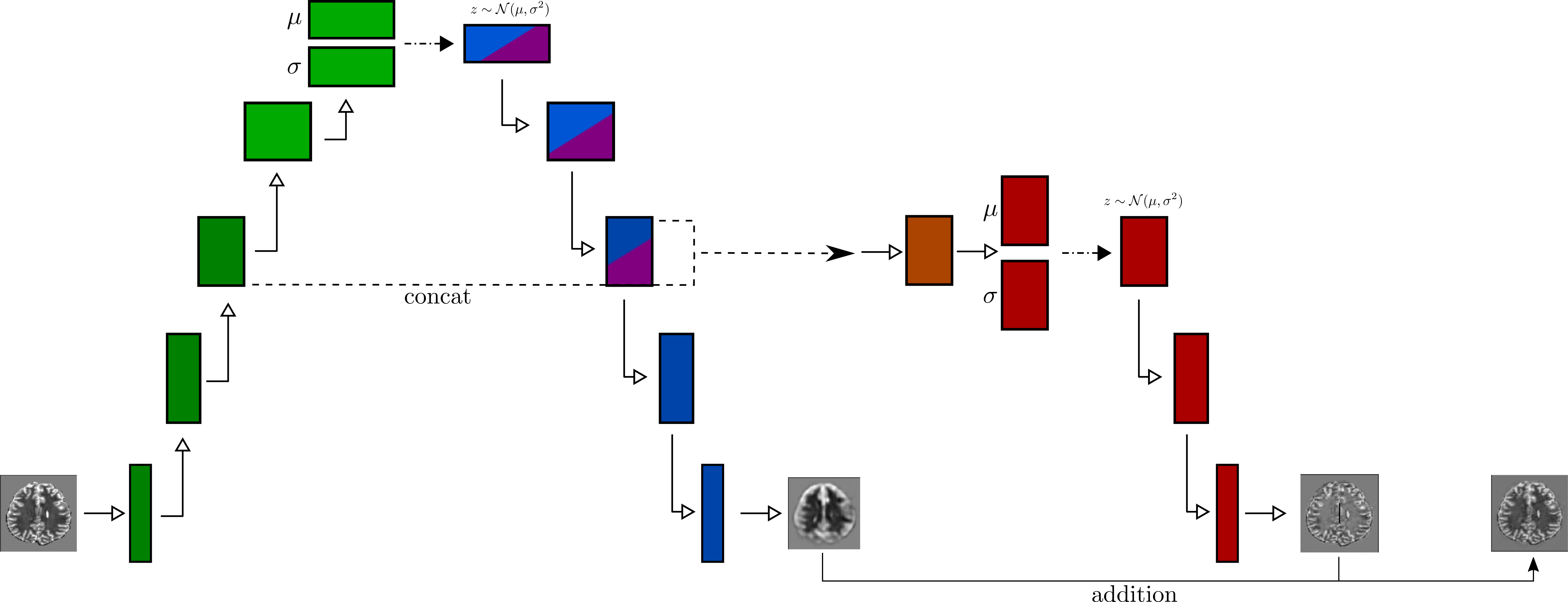}
  \caption{The pchVAE model is shown. The green part is the encoder, consisting of the low-level encoder $f_1$ (dark green) and the high-level encoder $f_2$ (lighter green). The blue part is the high-level decoder $g_1$ and the red part is the low-level decoder $g_2$. The arrows with straight lines and white heads represent (transposed) Conv-Ops, where the up-/down-ward convs have an up-/down-sampling factor of 2. }
  \label{fig:model}
\end{figure}

Recent research has show that VAEs pursue PCA directions (by accident) \cite{rolinek_variational_2018}. Inspired by this and the iterative PCA algorithm \cite{bro_principal_2014}, we propose the optimization problem for a linear AE with two components and extend it to a non-linear VAE (however in contrast to PCA which maximizes the variance for the components, this maximizes the mutual information of the input and the first component, see Suppl.). Given the following optimization problem for a linear AE with input $X$ and weights $w1, w2$:


\begin{equation}
    \min_{w_1,w_2} \lambda_1 || X - w_1 w_1^\top X || +  \lambda_2 || (X - w_1 w_1^\top X) - w_2 w_2^\top (X - w_1 w_1^\top X) || ,
    \label{eq:1}
\end{equation}

we derive a similar optimization problem: 
\begin{equation}
    \min_{w_1,w_2} \lambda_1 || X - w_1 w_1^\top X || +  \lambda_2 || X - (w_1 w_1^\top X + w_2 w_2^\top X ) ||  + \lambda_3 ||   w_2 w_2^\top w_1 w_1^\top X || .
    \label{eq:2}
\end{equation}
Instead of using linear functions/ multiplications, replacing the weight matrices with arbitrary non-linear functions parameterized by neural networks and by amortizing the optimization over mini-batches/samples, gives the following optimization problem: 
\begin{equation}
\begin{split}
    \min_{\theta_1 , \theta_2, \gamma_1, \gamma_2} \lambda_1 || x - g_{\theta_1}(f_{\gamma_1}(x)) || +  \lambda_2 || x - (g_{\theta_1}(f_{\gamma_1}(x)) + g_{\theta_2}(f_{\gamma_1}(x), f_{\gamma_2}(x)) )|| \\ + \lambda_3 ||   g_{\theta_1}(f_{\gamma_1}(g_{\theta_2}(f_{\gamma_1}(x), f_{\gamma_2}(x))) || .
    \label{eq:3}
\end{split}
\end{equation}

where $f_{\gamma_1}, f_{\gamma_2}$ are the encoders (sharing the first layers) and $g_{\theta_1}, g_{\theta_2}$ are the decoders. From a information theoretical point the last term be interpreted as minimizing the mutual information between the low-level components and the high-level components \cite{chen_infogan:_nodate}.
We integrate this formulation in a conditional hierarchical VAE with a similar architecture to VQVAE-2, by using a normal prior for the latent variables $z_1$, $z_2$ and condition $ g_{\theta_2}$ not just on $z_2$ ($ \sim f_{\gamma_2}(x)$) but also on $z_1$ ($ \sim f_{\gamma_1}(x)$).
An overview of the complete architecture can be seen in Fig. \ref{fig:model}. 

\section{Experiments \& Results}

To evaluate the performance we consider two different aspects, first the reconstruction performance on a held-back test set and secondly the semantic OoD \cite{ahmed_detecting_2019} / anomaly detection performance. The OoD performance can hopefully give truthful insights into the capturing of the data distribution, since likelihood-based comparisons have often been criticized \cite{nalisnick_deep_2018, theis_note_nodate}.

The models are trained on 2D slices of 800 brain MRIs from the HCP dataset \cite{van_essen_human_2012}. 
For the reconstruction performance, we show the MSE for 200 held-back patients from the HCP dataset.
For the OoD/ anomaly detection we evaluate the performance on two datasets. As first dataset we use the 200 held-back patients from the HCP dataset and randomly render natural objects in the brain area of some slices (similar to \cite{drew_invisible_2013}), which are then considered anomalies.  As a second dataset we use the BraTS2017 \cite{bakas_advancing_2017, menze_multimodal_2015} dataset where slices with tumor annotations are considered anomalies. 
The approximated ELBO is chosen as OoD/ anomaly score for each slice, as is common practice  \cite{an_variational_2015,kiran_overview_2018}.  
We report the mean area under the receiver operator curve (AUROC) and average precision (AP, as suggested by \cite{ahmed_detecting_2019}).
We compare the pchVAE with different other VAE models with slightly more parameters and the same number of latent variables (if possible). 
The first VAE variant, the ``high-level'' VAE (High VAE) is similar to the high-level component branch of the pchVAE (the green and blue parts in Fig. \ref{fig:model}) with 4 up-sampling operations. A second variant, the ``low-level'' VAE (Low VAE) only consists of the low-level part of the pchVAE model (the dark green and dark blue parts in Fig. \ref{fig:model}) with 2 up-sampling operations. We also consider a default conditional hierarchical VAE (chVAE) (similar to the model in Fig \ref{fig:model} but without the dark blue part and term 3 in Eq. \ref{eq:3}) and the VQVAE-2 (\cite{noauthor_vqvae-2_nodate}). Each model was trained for 10 epochs (which based on a validation set showed convergence for all models) with Adam with a learning-rate of $0.0001$ and a batch-size of 64. 
 
\begin{figure}
  \centering
    \includegraphics[width=0.9\textwidth]{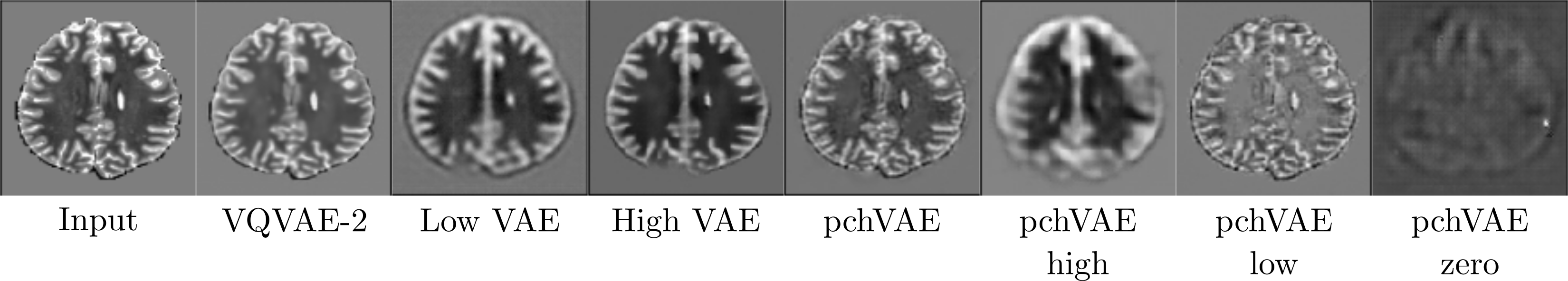}
  \caption{Different reconstructions of the same input. Reconstructions for the VQVAE-2, the Low VAE, the High VAE and the pchVAE are shown. Furthermore for reconstruction of pchVAE is furthermore spilt in the high-level components (Eq. \ref{eq:3} term 1) and the low-level components (Eq. \ref{eq:3} term 2) and the low-level components passed through the high-level encoder \& decoder is given as pchVAE zero (Eq. \ref{eq:3} term 3).}
  \label{fig:imgs}
\end{figure}

The results (Table \ref{sample-table} and Fig. \ref{fig:imgs}) show a similar or slightly better performance of the pchVAE compared to the High VAE on the OoD task, with the pchVAE having superior reconstruction performance. However, on the reconstruction task, the Low VAE and VQVAE-2 show a slightly better performance. These two model have a worse OoD performance indicating that those models have focused more low-level statistics of the data and in this case, often fail to capture the data distribution \cite{nalisnick_deep_2018}.

\begin{table}
\centering

\begin{tabular}{c|c|cc|}
\cline{2-4}
         & Reconstruction & \multicolumn{2}{c|}{OoD} \\ \cline{2-4} 
         & MSE $\downarrow{}$	& AUROC $\uparrow{}$   & AP $\uparrow{}$  \\ \hline
VQVAE-2  & $0.02823 \pm 0.0001$            & $0.6965 \pm 0.0036 $    & $0.4585 \pm 0.0058$  \\
Low VAE  & $\textbf{0.01169} \pm 0.0002$            & $0.7101 \pm 0.0010$    & $0.4922 \pm 0.0025$  \\
High VAE & $0.07036 \pm 0.0007$  & $0.7207 \pm 0.0005$    & $0.5154 \pm 0.0010$  \\
chVAE    & $0.02124 \pm 0.0040$  & $0.6716 \pm 0.0009$    & $0.4341 \pm 0.0008$  \\
\textit{pchVAE}   & $0.03224 \pm 0.0017$           & $\textbf{0.7277} \pm 0.0002$    & $\textbf{0.5321} \pm 0.0017$ \\ \hline
\end{tabular}
  \caption{Mean and standard deviation of reconstruction and OoD performance of the different models over five different runs.}
  \label{sample-table}
\end{table}



\section{Discussion \& Conclusion}

The results show a promising performance of the pchVAE. Adding a conditioned low-level reconstruction branch and an additional forward pass to a VAE allows us to trade off computational overhead for a better reconstruction performance. We think especially for medical anomaly detection and localization tasks a good reconstruction performance with a focus to details is important.  

Other approaches such as adversarial losses \cite{baur_deep_2018, larsen_autoencoding_2016} have also aimed at more truthful and sharper reconstructions. However, while these adversarial losses lead to sharper reconstructions due to their nature they can not create more low-level information and thus often ``dream-up'' plausible information which leads to a higher MSE and can for medical applications potentially be dangerous. We did not yet compare our model with BIVA but think that due to the simpler nature of our model it holds some benefits.

Here we demonstrated a way to improve the reconstruction performance of VAEs while showing similar or slightly better anomaly detection performance. We think that this can represent an important step for unsupervised learning of data distributions for pretraining, feature extraction or anomaly detection.

\bibliography{references}
\bibliographystyle{abbrv}

\newpage
\section{Supplements}

\subsection{Mutual Information between the input and encoding}

Given the input $x$, an encoder $f$, a decoder $g$ and the encoding $z$, we can, similar to \cite{chen_infogan:_nodate}, show the relations between optimizing the reconstruction error of an AE and the mutual information as :

\begin{equation}
\begin{split}
    \mathcal{I}(x;F(x)) & = H(x) - H(x|F(x)) \\
     & = \mathbb{E}_{z \sim F(x)}[\mathbb{E}_{x' \sim P(x|z)} [\log P(x'|z)]] + H(x) \\
     & =  \mathbb{E}_{z \sim F(x)}[D_{KL}(P(\cdot | x) || G(\cdot | x)) + \mathbb{E}_{x' \sim P(x|z)} [\log G(x'|z)]] + H(x) \\
     & \ge \mathbb{E}_{z \sim F(x)}[\mathbb{E}_{x' \sim P(x|z)} [\log G(x'|z)]] + H(x),
\end{split}
\end{equation}
where  $P(x|z)$ is the intractable generative distribution and we use an auxiliary distribution $G(x|z)$ to approximate $P(x|z)$. Choosing $G(x|z)$ as a normal distribution with a constant variance and the mean parameterized by a neural network $g$, can collapse it back to an neural network decoder with a MSE loss. Furthermore $H(x)$ is given by the data distribution and can be assumed to be constant. 

Using Lemma A. 1 from \cite{chen_infogan:_nodate} can simplify the equation and eliminate the intractable sampling from $x' \sim P(x|z)$: 

\begin{equation}
\begin{split}
    \mathcal{I}(x;F(x)) & = H(x) - H(x|F(x)) \\
     & \ge \mathbb{E}_{z \sim F(x)}[\mathbb{E}_{x' \sim P(x|z)} [\log G(x'|z)]] + H(x)
     \\
     & =  \mathbb{E}_{x \sim P(x), z \sim F(x)}[\log G(x|z)] ] + H(x),
\end{split}
\end{equation}

which now in analogy to VAEs can be solved with the reparametrization trick and Monte Carlo simulation where $F$ is the inference/ encoding distribution parameterized by a neural network $f$. 

\subsection{Further clarification of the loss function}

Next we will further clarify the loss given in Eq. (1)-(3) and show the resulting loss function for the pchVAE.

\begin{equation}
\begin{split}
   & \lambda_1 || X - w_1 w_1^\top X || +  \lambda_2 || (X - w_1  w_1^\top X) - w_2 w_2^\top (X - w_1 w_1^\top X) || \\
   = & \lambda_1 || X - w_1 w_1^\top X || +  \lambda_2 || X - w_1  w_1^\top X - w_2 w_2^\top X + w_2 w_2^\top w_1 w_1^\top X) || \\ 
   \stackrel{\mathclap{\normalfont\mbox{(triangle inequality)}}}{\le} 
   & \lambda_1 || X - w_1 w_1^\top X || +  \lambda_2 || X - w_1  w_1^\top X - w_2 w_2^\top X || + \lambda_2 || w_2 w_2^\top w_1 w_1^\top X) ||\\
   = & 
    \lambda_1 || X - w_1 w_1^\top X || +  \lambda_2 || X - (w_1 w_1^\top X + w_2 w_2^\top X ) ||  + \lambda_2 ||   w_2 w_2^\top w_1 w_1^\top X || .
\end{split}
\end{equation}

For further modularity we introduce $\lambda_3$ instead of $\lambda_2$ for the third term. However, in practice we chose $\lambda_2 = \lambda_3$ .

To take the step from a linear completely symmetric AE to a non-linear AE we substitute the matrix multiplications with arbitrary functions approximated by neural networks: 

\begin{equation*}
\begin{split}
w_1 w_1^\top x & \xrightarrow{becomes} g_{\theta_1}(f_{\gamma_1}(x)), \\
w_2 w_2^\top X & \xrightarrow{becomes} g_{\theta_2}(f_{\gamma_1}(x), f_{\gamma_2}(x)), \\
w_2 w_2^\top w_1 w_1^\top & \xrightarrow{becomes} g_{\theta_1}(f_{\gamma_1}(g_{\theta_2}(f_{\gamma_1}(x), f_{\gamma_2}(x))) . 
\end{split}
\end{equation*}

This results in the following formulation: 
\begin{equation}
 \lambda_1 || x - g_{\theta_1}(f_{\gamma_1}(x)) || +  \lambda_2 || x - (g_{\theta_1}(f_{\gamma_1}(x)) + g_{\theta_2}(f_{\gamma_1}(x), f_{\gamma_2}(x)) )|| \\ + \lambda_3 ||   g_{\theta_1}(f_{\gamma_1}(g_{\theta_2}(f_{\gamma_1}(x), f_{\gamma_2}(x))) || .
\end{equation}

However, since we are interested to find the latent factors of a generative hierarchical model we, in analogy to \cite{kingma_auto-encoding_2013, rezende_stochastic_2014}, integrate this is into the variational auto-encoding framework:
\begin{equation}
\begin{split}
 L = &
 \lambda_1  \mathbb{E}_{z_1 \sim \mathcal{N}(f_{\gamma_1, \mu}(x), f_{\gamma_1, \sigma}(x))} \mathcal{N}(x;g_{\theta_1}(z_1),c)  \\ + & D_{KL}(\mathcal{N}(f_{\gamma_1, \mu}(x), f_{\gamma_1, \sigma}(x)) || \mathcal{N}(0,1) ) \\ + &
 \lambda_2 \mathbb{E}_{z_1 \sim \mathcal{N}(f_{\gamma_1, \mu}(x), f_{\gamma_1, \sigma}(x))} \mathbb{E}_{z_2 \sim \mathcal{N}(f_{\gamma_2, \mu}(x, z_1), f_{\gamma_2, \sigma}(x, z_1))} \mathcal{N}(x;g_{\theta_1}(z_1)+g_{\theta_2}(z_1, z_2),c) \\ + &
  D_{KL}(\mathcal{N}(f_{\gamma_2, \mu}(x, z_1), f_{\gamma_2, \sigma}(x, z_1)) || \mathcal{N}(0,1) ) \\ + &
 \lambda_3 
 \mathbb{E}_{z_1 \sim \mathcal{N}(f_{\gamma_1, \mu}(x), f_{\gamma_1, \sigma}(x))}
 \mathbb{E}_{z_2 \sim \mathcal{N}(f_{\gamma_2, \mu}(x, z_1), f_{\gamma_2, \sigma}(x, z_1))}  \\ + &
 \mathbb{E}_{x_{low} \sim \mathcal{N}(g_{\theta_2}(z_1, z_2)}) 
 \mathbb{E}_{z_{low} \sim \mathcal{N}(f_{\gamma_1, \mu}(x_{low}), f_{\gamma_1, \sigma}(x_{low}))} \mathcal{N}(0;g_{\theta_1}(z_{low}),c),
\end{split}
\end{equation}
where the last term encourages the mutual information in the conditioned low-level component $x_low$ and the high level encoding to be zero. Using Monte Carlo sampling with sampling size 1, choosing $c$ appropriately and integrating the reparametrization step into the decoders $g$ gives the final and familiar loss function: 
\begin{equation}
\begin{split}
 L & = \lambda_1 || x - g_{\theta_1}(f_{\gamma_1}(x)) || +  \lambda_2 || x - (g_{\theta_1}(f_{\gamma_1}(x)) + g_{\theta_2}(f_{\gamma_1}(x), f_{\gamma_2}(x)) )|| \\ & + \lambda_3 ||   g_{\theta_1}(f_{\gamma_1}(g_{\theta_2}(f_{\gamma_1}(x), f_{\gamma_2}(x))) || \\ &
 + \lambda_4 D_{KL}(\mathcal{N}(f_{\gamma_1, \mu}(x), f_{\gamma_1, \sigma}(x)) || \mathcal{N}(0,1) )  \\ & + \lambda_5 D_{KL}(\mathcal{N}(f_{\gamma_2, \mu}(x, f_{\gamma_1}(x)), f_{\gamma_2, \sigma}(x, f_{\gamma_1}(x))).
\end{split}
\end{equation}

\newpage

\subsection{Reconstructions from different models}

\begin{figure}[h!]
    \centering
    \begin{subfigure}[b]{0.4\textwidth}
        \includegraphics[width=\textwidth]{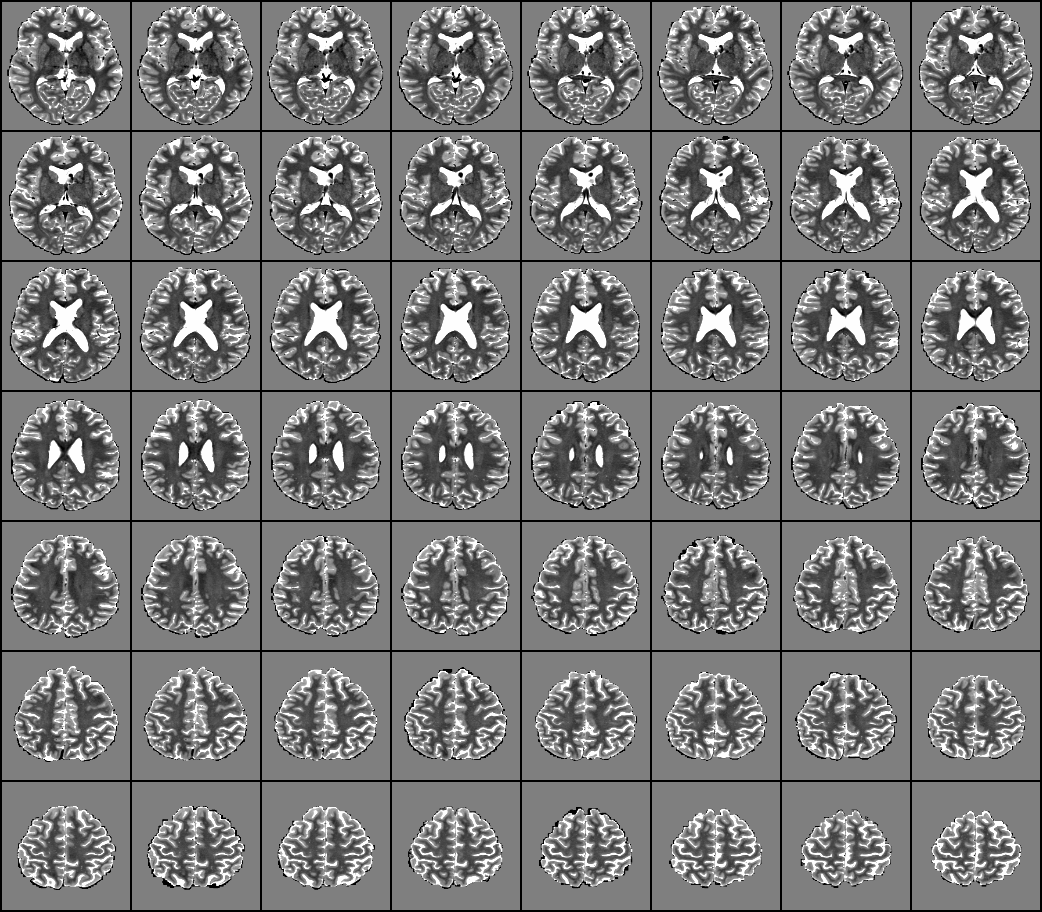}
        \caption{Input}
    \end{subfigure}
    \begin{subfigure}[b]{0.4\textwidth}
        \includegraphics[width=\textwidth]{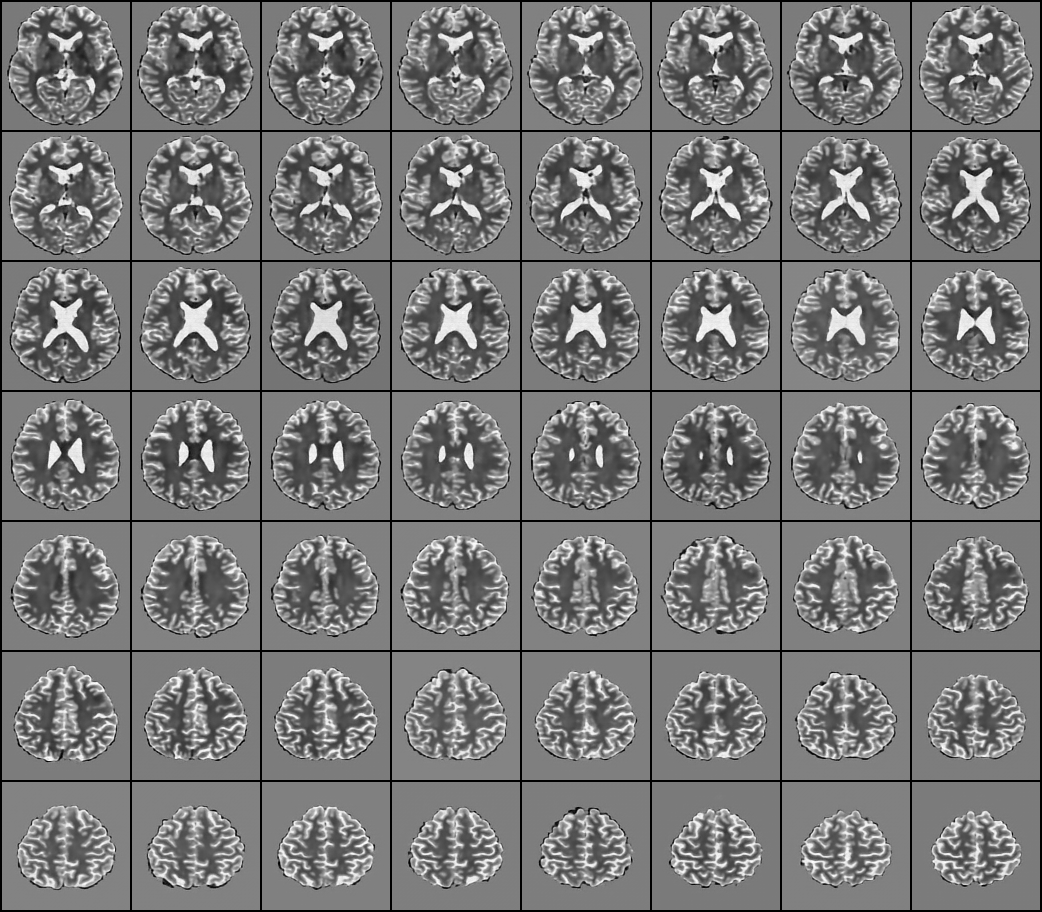}
        \caption{VQVAE-2 reconstruction}
    \end{subfigure}
    
    \begin{subfigure}[b]{0.4\textwidth}
        \includegraphics[width=\textwidth]{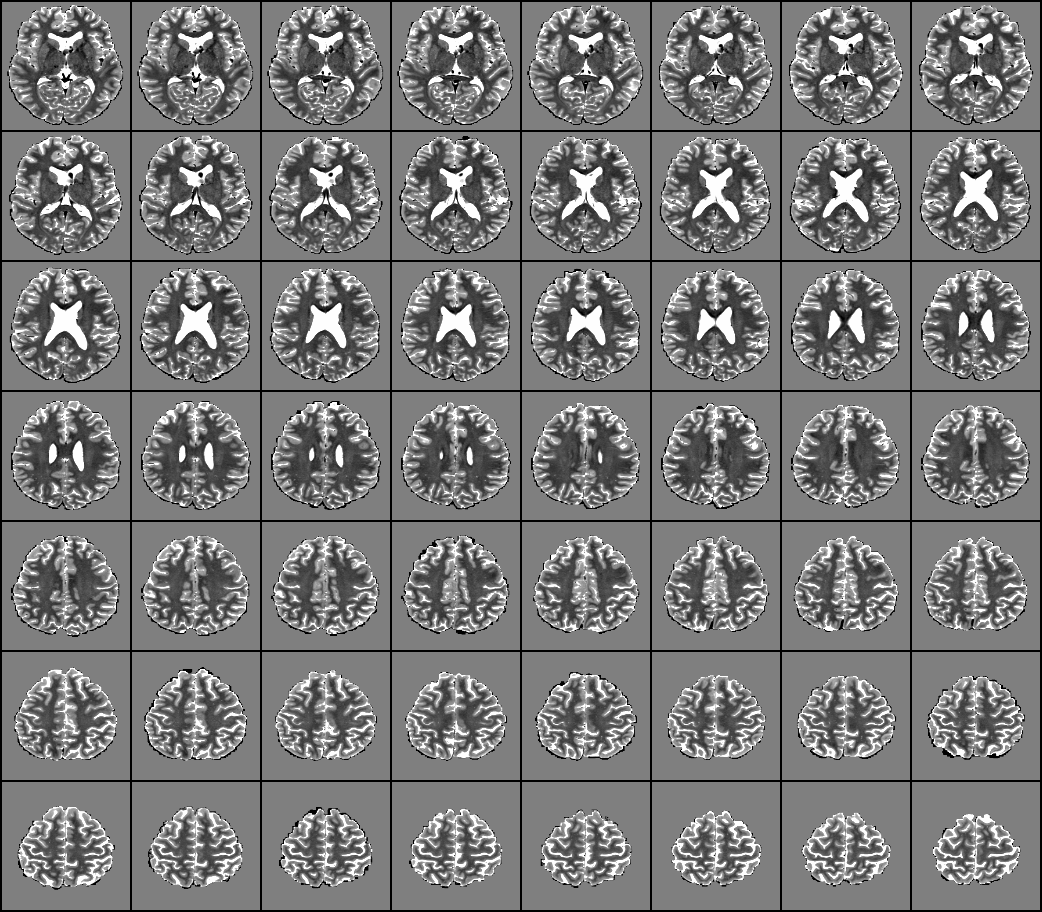}
        \caption{Input}
    \end{subfigure}
    \begin{subfigure}[b]{0.4\textwidth}
        \includegraphics[width=\textwidth]{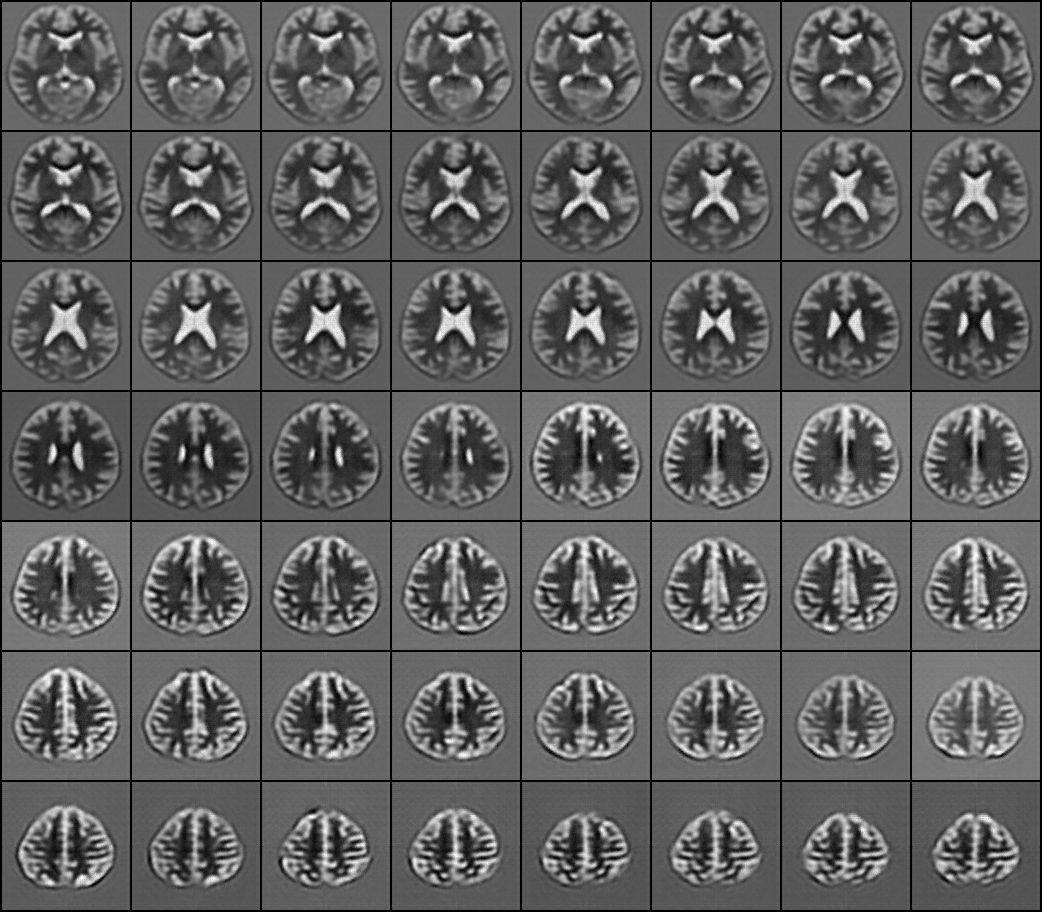}
        \caption{VAE-low reconstruction}
    \end{subfigure}
    
    \begin{subfigure}[b]{0.4\textwidth}
        \includegraphics[width=\textwidth]{figs/reconstructions/input.png}
        \caption{Input}
    \end{subfigure}
    \begin{subfigure}[b]{0.4\textwidth}
        \includegraphics[width=\textwidth]{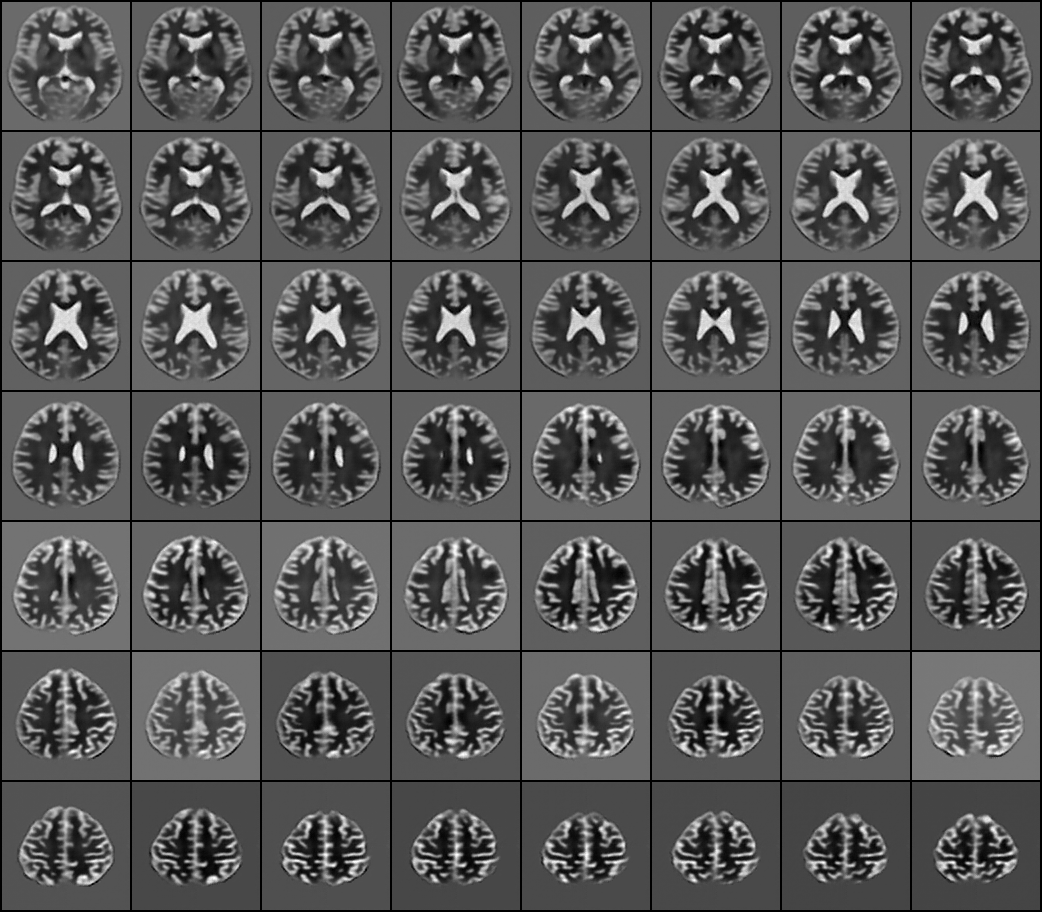}
        \caption{VAE-high reconstruction}
    \end{subfigure}
    
    \begin{subfigure}[b]{0.4\textwidth}
        \includegraphics[width=\textwidth]{figs/reconstructions/input.png}
        \caption{Input}
    \end{subfigure}
    \begin{subfigure}[b]{0.4\textwidth}
        \includegraphics[width=\textwidth]{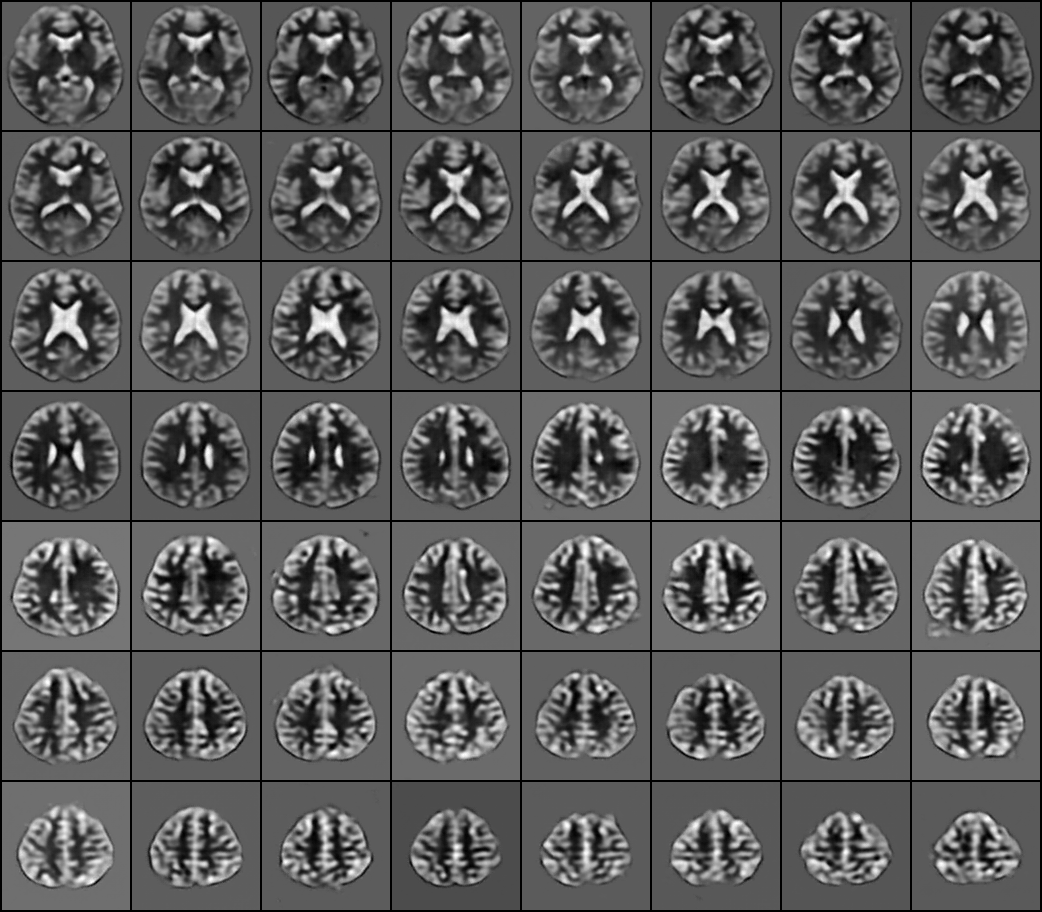}
        \caption{pchVAE reconstruction}
    \end{subfigure}

\end{figure}

\newpage
 
\subsection{pchVAE reconstructions}

\begin{figure}[h!]
    \centering
    \begin{subfigure}[b]{0.4\textwidth}
        \includegraphics[width=\textwidth]{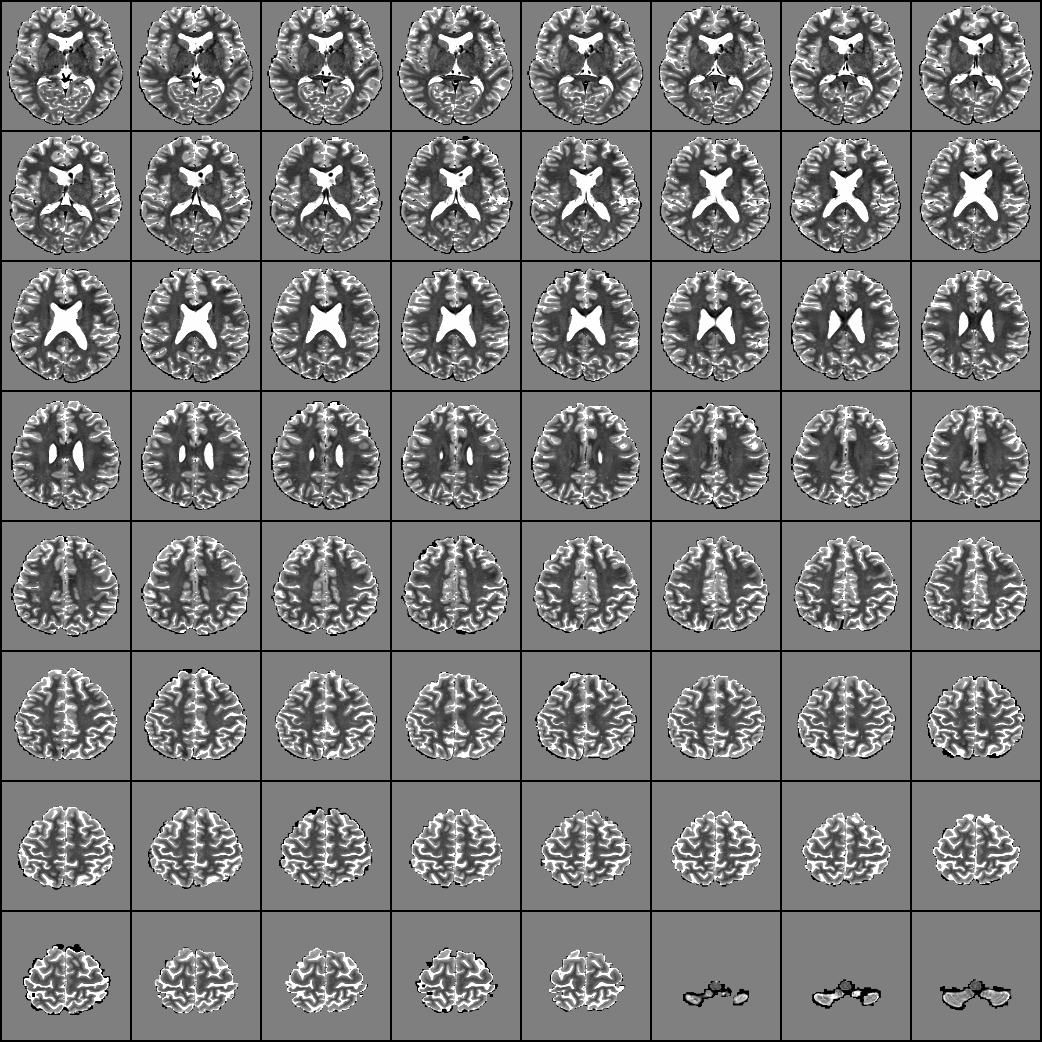}
        \caption{Input}
    \end{subfigure}
    \begin{subfigure}[b]{0.4\textwidth}
        \includegraphics[width=\textwidth]{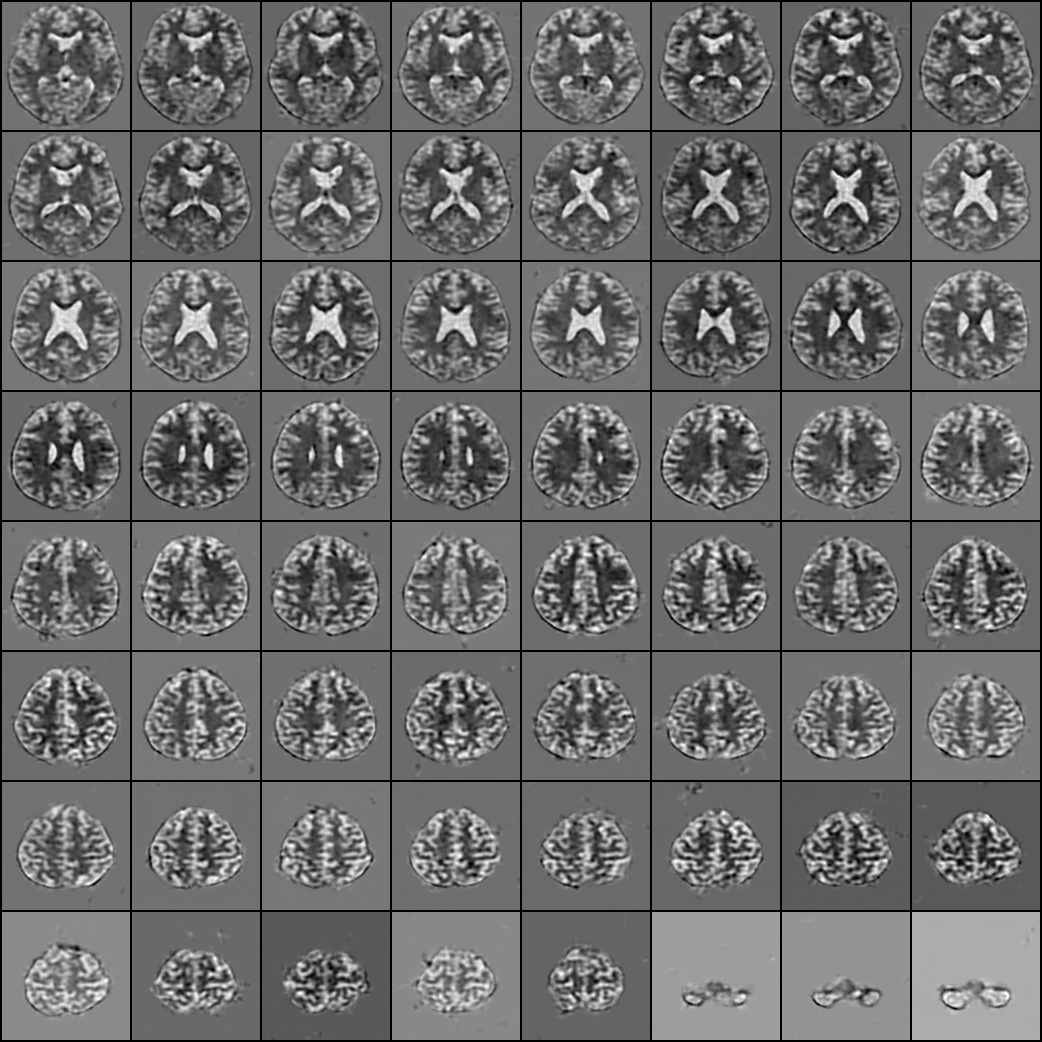}
        \caption{Complete pchVAE reconstruction}
    \end{subfigure}
    \begin{subfigure}[b]{0.4\textwidth}
        \includegraphics[width=\textwidth]{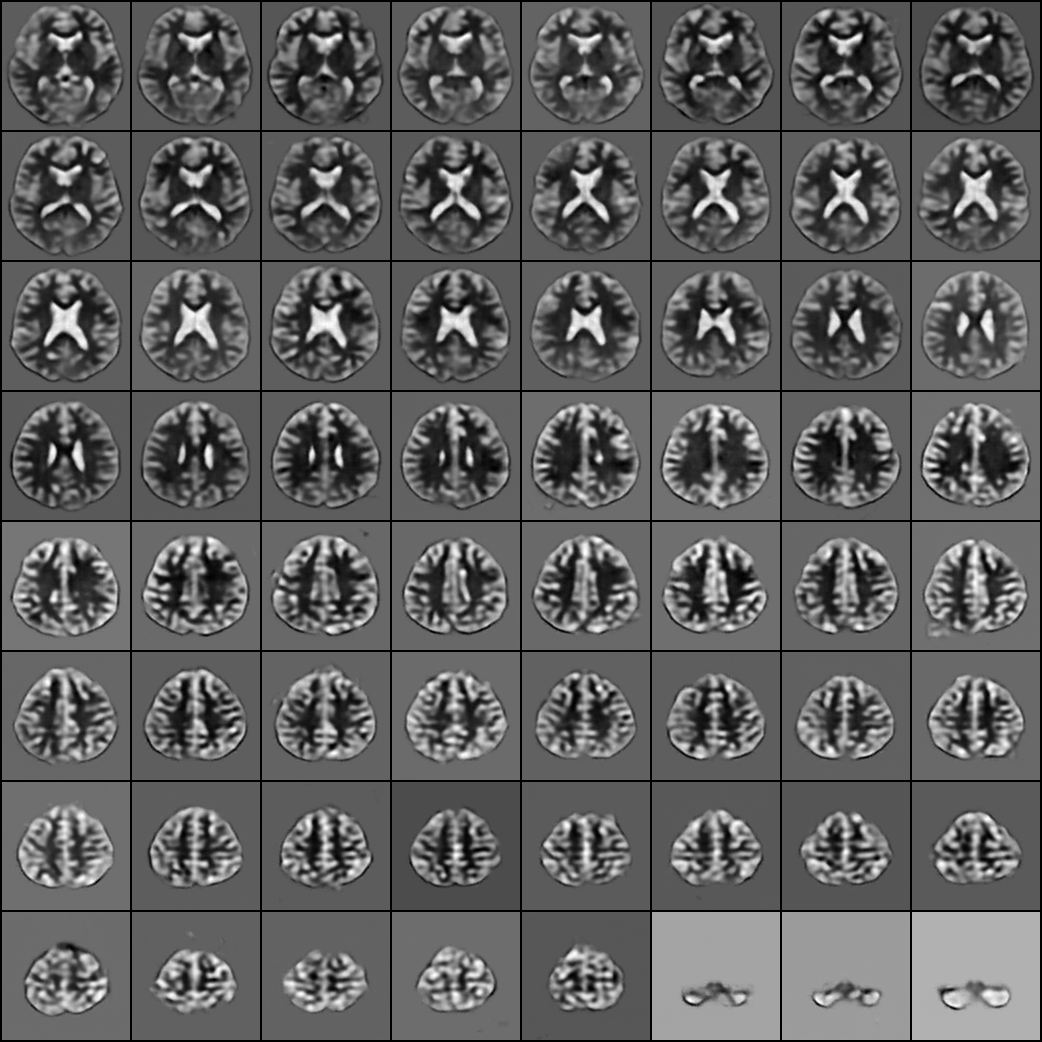}
        \caption{pchVAE high (Eq. 3 term 1) }
    \end{subfigure}
    \begin{subfigure}[b]{0.4\textwidth}
        \includegraphics[width=\textwidth]{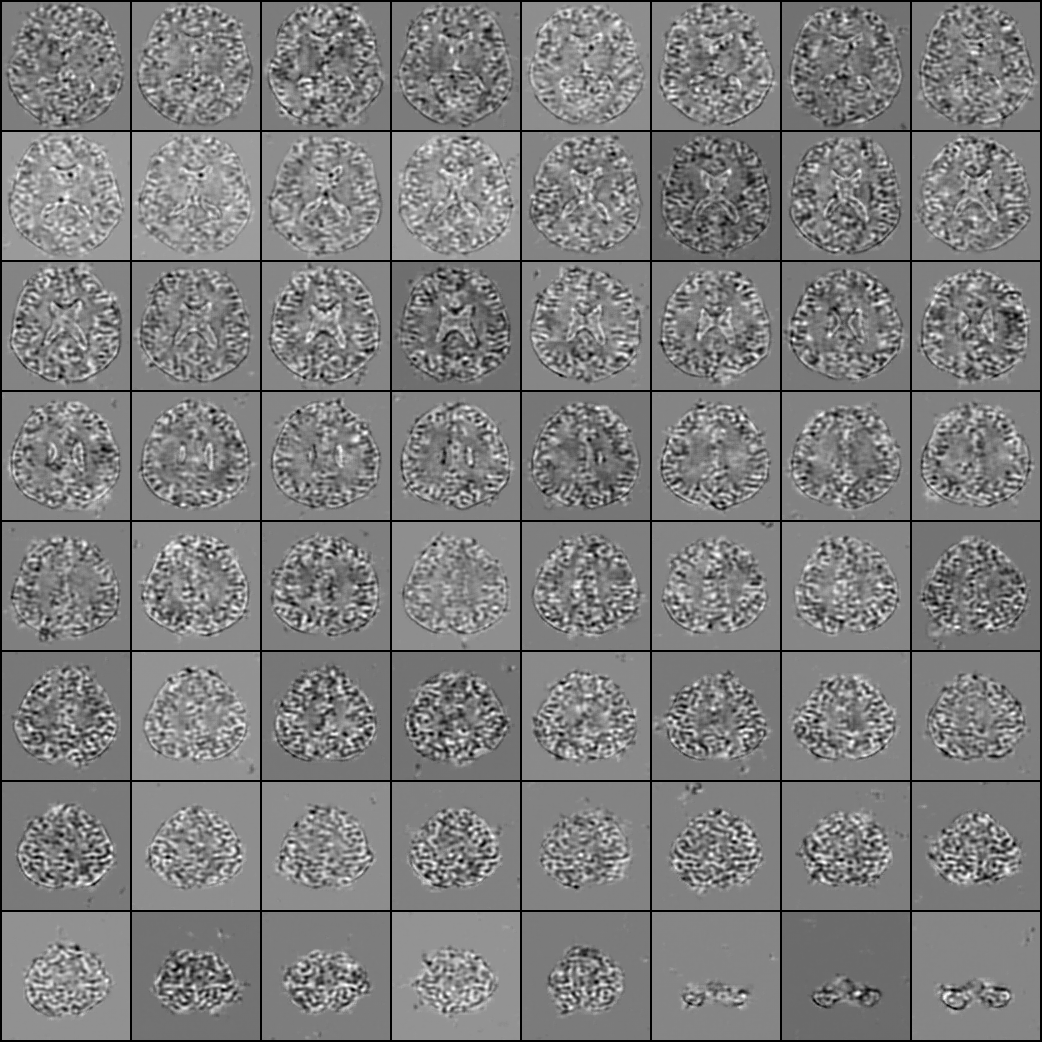}
        \caption{pchVAE low (Eq. 3 term 2) }
    \end{subfigure}
    \begin{subfigure}[b]{0.4\textwidth}
        \includegraphics[width=\textwidth]{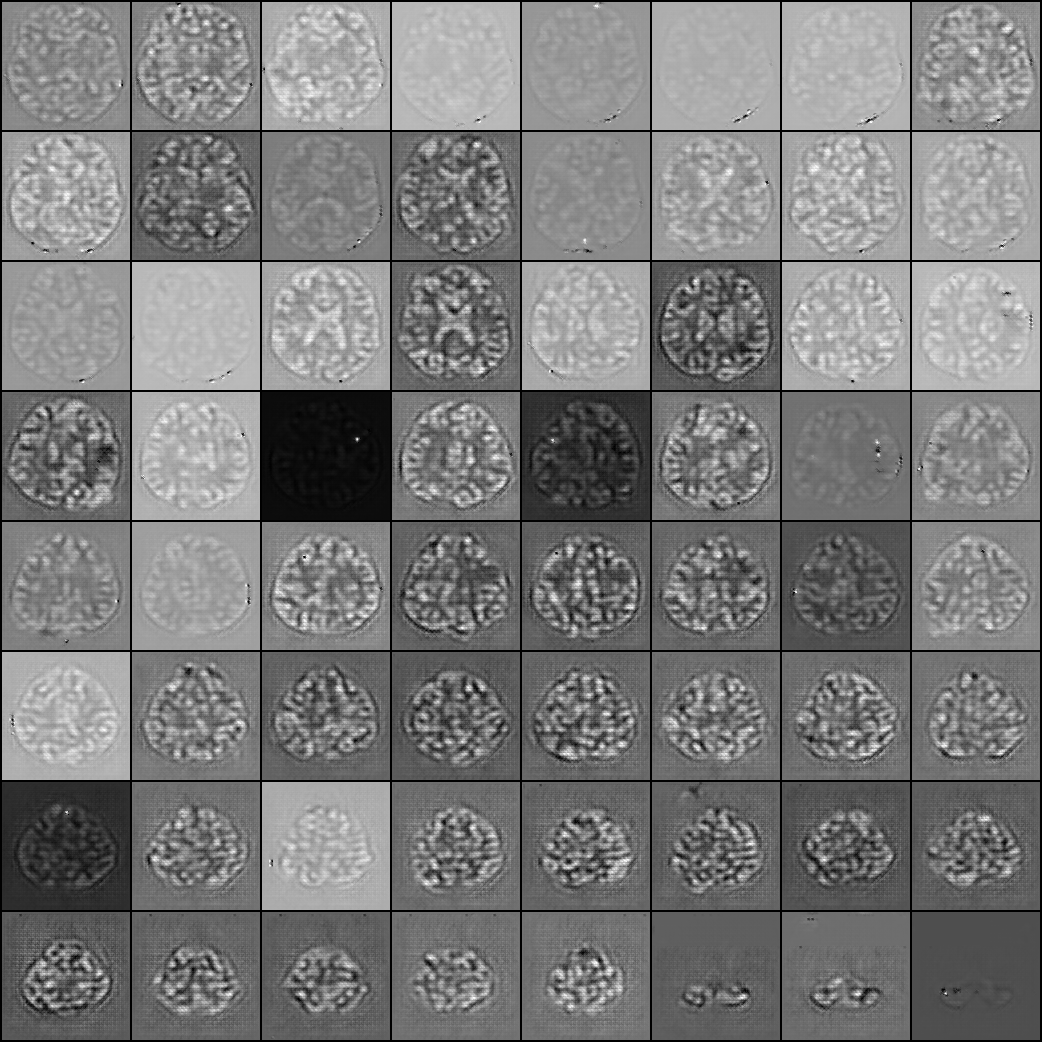}
        \caption{pchVAE zero (Eq. 3 term 3) }    \end{subfigure}

\end{figure}

\end{document}